\documentclass[sigconf, nonacm]{acmart}

\usepackage[ruled,linesnumbered]{algorithm2e}
\usepackage{courier}
\usepackage{multirow}
\usepackage{url}

\AtBeginDocument{%
  \providecommand\BibTeX{{%
    \normalfont B\kern-0.5em{\scshape i\kern-0.25em b}\kern-0.8em\TeX}}}

\begin{document}

\title{Towards Learning Instantiated Logical Rules from Knowledge Graphs}

\author{Yulong Gu}
\affiliation{%
  \institution{Newcastle University}
  \country{United Kingdom}
}
\email{y.gu11@newcastle.ac.uk}

\author{Yu Guan}
\affiliation{%
  \institution{Newcastle University}
  \country{United Kingdom}
}
\email{yu.guan@newcastle.ac.uk}

\author{Paolo Missier}
\affiliation{%
  \institution{Newcastle University}
  \country{United Kingdom}
}
\email{paolo.missier@newcastle.ac.uk}


\begin{abstract}
Efficiently inducing high-level interpretable regularities from knowledge graphs (KGs) is an essential yet challenging task that benefits many downstream applications. In this work, we present GPFL, a probabilistic rule learner optimized to mine instantiated first-order logic rules from KGs. Instantiated rules contain constants extracted from KGs. Compared to abstract rules that contain no constants, instantiated rules are capable of explaining and expressing concepts in more details. GPFL utilizes a novel two-stage rule generation mechanism that first generalizes extracted paths into templates that are acyclic abstract rules until a certain degree of template saturation is achieved, then specializes the generated templates into instantiated rules. Unlike existing works that ground every mined instantiated rule for evaluation, GPFL shares groundings between structurally similar rules for collective evaluation. Moreover, we reveal the presence of overfitting rules, their impact on the predictive performance, and the effectiveness of a simple validation method filtering out overfitting rules. Through extensive experiments on public benchmark datasets, we show that GPFL 1.) significantly reduces the runtime on evaluating instantiated rules; 2.) discovers much more quality instantiated rules than existing works; 3.) improves the predictive performance of learned rules by removing overfitting rules via validation; 4.) is competitive on knowledge graph completion task compared to state-of-the-art baselines.
\end{abstract}




\maketitle

\section{Introduction}
A Knowledge Graph (KG) is a graph-based abstraction 
of knowledge where entities are represented by nodes and facts by relationships
between nodes \cite{Fensel2020}.For instance, the fact "Beijing is the capital of China" can be
represented as a relationship \textit{Capital\_of(Beijing, China)} where Beijing is
the subject, Capital\_of the predicate, and China the object of the relationship.
KGs intuitively represent domains that involve interactions between entities, such as 
social relationships, biological interactions and bibliographical citations. Over the last 
decade, many large KGs have been created, including NELL \cite{Mitchell2018}, Freebase \cite{Bollacker2008} and DBpedia \cite{Auer2007}, to support intelligent applications. 
Reasoning over KGs aims to reveal implicit knowledge through the understanding of 
existing facts. Among many approaches that reason over KGs \cite{Chen2020}, rule learning 
methods \cite{Galarraga2015, Omran2018, Meilicke2019} that generate 
first-order logic rules based on ontological and relational information that present in KGs 
have attracted increasing attentions for being inductive, interpretable and transferable. 
The successful application of first-order logic rules to fact checking \cite{Gad-Elrab2019}, 
question answering \cite{Zhang2018} and knowledge graph completion \cite{Meilicke2018} 
has demonstrated promising potential of utilizing rule learners for various 
downstream tasks.

One differentiating factor that implicitly categorizes rule learners is the types
of rules a learner produces. For instance, QuickFOIL \cite{Zeng2014} and ScaLeKB \cite{Chen2016}
produce Horn rules for deducing unknown facts; RuLES \cite{Ho2018} learns non-monotonic rules for
exception handling; RuDiK \cite{Ortona2018} proposes negative rules to identify contradictions in
the data, and AMIE+ \cite{Galarraga2015} and AnyBURL \cite{Meilicke2019} include instantiated rules
that contain constants to enrich the expressivity of the learned rule space. In this work, we
focus on developing rule learners that mine probabilistic positive Horn rules directly
from KGs in general, and are especially optimized at discovering instantiated rules for better 
expressivity.

\begin{figure}
\centerline{\includegraphics[scale=0.4]{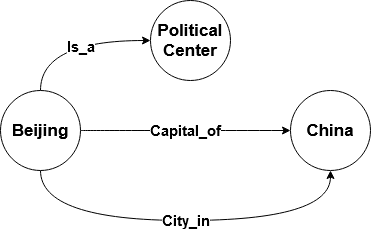}}
\caption{A small knowledge graph.}
\label{fig.1}
\end{figure}

\subsection{Challenge}

The rule space of a rule learner is a set containing all possible rules that can be produced
by the learner under various constraints. The balance between rule space complexity for model 
expressivity and scalability for system practicability remains one of the core challenges in 
rule learning. The complexity of a rule space is determined by the choice of language 
bias \cite{de2008logical} that dictates what types of rules to include, and different language 
biases often render rule spaces differing drastically in size.

Let us take the evolution of Path Ranking Algorithms (PRAs) as an example. PRA \cite{Lao2011} uses random 
walkers to extract a certain type of rules, the Closed Abstract Rules (CARs) that are
cyclic sequences of predicates connecting entity pairs. For instance in Figure.\ref{fig.1}, 
given predicate \textit{Capital\_of(X, Y)} as the learning target, and entity pair 
\textit{(Beijing, China)} as a positive instance, we can manually induce the CAR:
\begin{displaymath}
    f_0 = \text{\textit{Capital\_of(X, Y) $\leftarrow$ City\_in(X, Y)}}
\end{displaymath}
which translates as if a city $X$ is in a country $Y$, then $X$ is the capital of $Y$.
It is apparently too general to describe the idea about capital. To explain ideas in more details, 
Cor-PRA \cite{Lao2015} proposes to include Tail Anchored Rules (TARs), a type of instantiated rules with 
the last variable being substituted by a constant. Again in Figure.\ref{fig.1}, we can induce TAR: 
\begin{displaymath}
  f_1 = \text{\textit{Capital\_of(X, Y) $\leftarrow$ Is\_a(X, "Political Center")}}
\end{displaymath}
which contains constant "Political Center". Now the conjunction of $f_0$ and $f_1$ is translated as
if a city $X$ is in a country $Y$, and $X$ is a political center, then $X$ is the capital of $Y$.

Although the inclusion of instantiated rules comes with improved expressivity, it also exposes the system 
to a much larger rule space than only including abstract rules that contain no constants. For instance,
assume the cardinality of the object of predicate \textit{Is\_a(X, $V_1$)} is $m$, from a simple abstract rule:
\begin{displaymath}
  f_2 = \text{\textit{Capital\_of(X, Y) $\leftarrow$ Is\_a(X, $V_1$)}}
\end{displaymath}
that subsumes TAR $f_1$ with respect to generality, we can derive $m$ instantiated rules by replacing variable
$V_1$ with constants, which is a $m$ times growth in rule space size from one abstract rule. On large KGs with
millions of relationships, the scale of their rule space with instantiated rules included is inordinate for
greedy search. Therefore, optimization and approximation approaches for efficient rule generation and evaluation
are needed.

\subsubsection{Rule Generation}

Rule generation procedure dictates how the rule space is traversed and when to stop the exploration. A majority of 
existing works \cite{Lao2011, Galarraga2015, Meilicke2018} explore the entirety or randomly sampled sub-spaces of 
the rule space. As the rule space that includes instantiated rules is often enormous, it is either too expensive or 
infeasible to search the entire rule space. Also, instantiated rules that are mined from randomly sampled sub-spaces 
are often subject to locality. AnyBURL \cite{Meilicke2019} proposes a feedback-aware mechanism to control the progress of rule generation. 
Specifically, rules of length $n$ are mined in batches where rules learned in previous batches are considered as known rules, and if
the proportion of known rules in current batch is above a saturation threshold, the system either progresses to mine rules of
length $n + 1$ or terminates. Although the feedback-awareness of AnyBURL is desirable in that the search is exposed to the entire 
rule space to mitigate sampling bias and early-stops when frequent regularities are extracted, it is inefficient at generating 
instantiated rules because to reach the saturation for progress, it needs to repeatedly visit the same set of frequent rules until 
a few less frequent rules are discovered. The root of this overhead is the hardness of its saturation convergence in that 
the rule space over which the saturation is measured is too large.

\subsubsection{Rule Evaluation}

Rule evaluation procedure decides what rule quality measure to use and how to plan the rule evaluation executions. 
Most of the existing works employ statistical measures such as confidence and support \cite{Galarraga2015} to reflect rule quality. 
Statistical measures are costly to compute in that it requires systems to ground rules using a backward chaining algorithm that is exponentially complex. 
In spite of the inefficiency, a majority of existing works ground every mined rule individually for evaluation, which leads to the main scalability bottleneck. 
Recent works propose to incorporate embeddings into the measure of rule quality. RLvLR \cite{Omran2018} uses quality measures based on embedding similarities 
to score rules efficiently. The disadvantage of embedding-augmented methods is that the training of embedding models on large KGs itself is often not trivial.

\subsection{Approach}

In this work, we propose the Graph Path Feature Learning (GPFL) system, a novel probabilistic rule learner 
optimized to mine instantiated rules. We use the idea of templates that are acyclic abstract rules to optimize both 
rule generation and evaluation. Specifically, GPFL utilizes a two-stage rule generation mechanism. In the 
generalization stage, GPFL optimizes the AnyBURL feedback-aware mechanism by saturating the template space to create 
a set of frequent templates. As the template space is usually smaller in size by orders of magnitude than the 
complete rule space over which AnyBURL measures its saturation, template saturation is much easier to converge and thus mitigates the overhead. Inspired by the idea of query pack \cite{Blockeel2002}, in the specialization stage, 
GPFL makes optimized use of template groundings for both deriving and evaluating instantiated rules. 
More specifically, as instantiated rules derived from the same templates are structurally similar, 
instead of grounding every mined rule individually, GPFL only grounds templates and uses the groundings of the 
templates to evaluate the derived instantiated rules collectively. In such a way, GPFL significantly reduces the 
number of invocations of the expensive grounding procedure compared to existing works. These optimizations allow GPFL to learn instantiated rules much more efficient than existing approaches 
in terms of quality and quantity. Moreover, as instantiated rules are more specialized than abstract rules, 
they are more likely to overfit the training set. GPFL removes the overfitting rules via a simple validation 
method for better predictive performance.

\subsection{Contribution}

Our contributions can be summarized as 1.) we propose a novel probabilistic rule learner optimized to mine instantiated rules; 
2.) to the best of our knowledge, this is the first work that utilizes a two-stage generalization-specialization mechanism to generate rules from KGs; 
3.) this is the first work that studies overfitting instantiated rules; 4.) GPFL is also the first rule learner that is  implemented on a graph database, 5.) through extensive experiments over public benchmark datasets, we observe that GPFL: 
significantly reduces the runtime on evaluating instantiated rules; mines much more quality rules than AnyBURL in a fixed time frame; improves the predictive performance of learned rules by removing overfitting rules via validation, and has competitive performance on knowledge graph completion task 
in comparison to state-of-the-art baselines.

\section{Related Works}

In this section, we discuss existing rule learning systems in detail.
We first give a brief introduction to differentiable rule learners, and
then examine classic rule learners that explore discrete rule space using
subsumption operators and score rules with statistical measures.

Neural LP \cite{Yang2017} and DRUM \cite{Sadeghian2019} are differentiable
rule learners based on the differentiable logic formulation TensorLog \cite{Cohen2020}
where the traversals over a KG are formulated as sequences of matrix multiplications
and logical inferences are compiled into sequences of differentiable operations on
matrices. Although the robustness introduced by numerical optimizations is desirable,
due to the limitations on current architecture designs, existing systems can only learn
CARs, which limits their expressivity.

Based on the mechanism employed for exploring discrete rule spaces, we categorize 
classic rule learners into Specialization-only (Spec), Generalization-only (Gen) 
and Specialization-Generalization (Spec-Gen) learners. Spec learners are top-down learners
such as QuickFOIL \cite{Zeng2014}, ScaLeKB \cite{Chen2016} and AMIE+ \cite{Galarraga2015}
that generate rules by repeatedly specializing rules derived from a top rule via adding
new atoms or instantiating variables in existing atoms. This approach often proposes
groundless rules that invoke the grounding procedure but do not contribute to the
inference. Gen learners are bottom-up learners such as PRA \cite{Lao2015}, SFE
\cite{Gardner2015}, RuleN \cite{Meilicke2018} and AnyBURL \cite{Meilicke2019} that
populate a rule space with rules generalized from paths in KGs. Compared to Spec methods,
Gen methods guarantee that there is at least one grounding for each generated rule. As it
is often impractical to extract all possible paths on large KGs, approximation strategies
are employed for scalability purpose. Spec-Gen learners include classic Inductive Logic
Programming (ILP) \cite{Muggleton2012} learners such as Progol \cite{Muggleton1995} and
Aleph \cite{srinivasan2001aleph} that first specialize the top rule into a bottom clause
which is a compact representation of a positive instance, and then generate rules by
generalizing bottom clauses.

In this work, we propose a Generalization-Specialization (Gen-Spec) mechanism that first
generalizes paths into templates and then specializes templates into instantiated rules. 
The collective evaluation strategy can efficiently identify groundless
rules, and the template saturation as an approximation strategy can effectively 
extract frequent high-level regularities without the need of searching the rule space exhaustively.

\section{Methodology}

A knowledge graph $\mathcal{G}=(\mathcal{E}, \mathcal{R}, \mathcal{T})$ is a directed multi-graph that contains ground atoms (facts) in the form of triples $r_i(e_j, e_k) \in \mathcal{T}$ where $r_i \in \mathcal{R}$ is a relationship type and $e_j, e_k \in \mathcal{E}$ are entities. In logic term, relationship type and entity are called predicate and constant, respectively. A path or ground rule, denoted by:
\begin{displaymath}
  r_0(e_0, e_1),r_1(e_1,e_2),...,r_n(e_n, e_{n+1})
\end{displaymath}
is a sequence of ground atoms extracted from KGs. The head of a rule is the first atom in the rule, and the rest of atoms are the body atoms. The length of a rule is the count of body atoms. We can rewrite a rule in definite Horn clause form as:
\begin{displaymath}
  r_0(e_0, e_1) \leftarrow r_1(e_1, e_2),...,r_n(e_n, e_{n+1})
\end{displaymath}
where the head atom $r_0(e_0, e_1)$ is inferred as a prediction for a KG, if all of the body atoms can be grounded in the KG. By replacing constants in ground atoms with variables, we can generalize ground rules into first-order logic rules.

\subsection{Language Bias}
Language bias, as a prior knowledge along with semantic bias, is used extensively in rule learning
methods to restrict rule space by specifying the desired types of rules to include
\cite{de2008logical}. For rule learners that generate rules based on paths extracted from KGs,
the implicit syntactic restrictions are that only binary atoms are allowed, and adjacent atoms
are connected by the same variables or constants. We use lower-case letters for constants and upper-case letters for variables where symbols $X$ and $Y$ denote variables in the head atom, and $V_i$ a variable in the body atoms. In this work, we only consider straight rules where a variable or constant can occur at most twice in the body atoms to avoid cycles. Also, we do not generate trivial rules that self-loop, such as:
\begin{displaymath}
  r_t(X, Y) \leftarrow r_1(X, V_0),..., r_n(V_n, X)
\end{displaymath}

Now we introduce some of the terms used throughout this work. Given a rule:
\begin{displaymath}
  r_t(X, Y) \leftarrow r_1(X, V_0),r_2(V_0, V_1),...,r_n(V_n, V_{n+1})
\end{displaymath}
we call variable $X$ the original variable in that the body atoms are originated from it; the variable $Y$ the free variable; variables such as $V_0$ the connecting variables in that they connect adjacent atoms, and the non-connecting variable $V_{n+1}$ in the last body atom the tail variable. A rule is closed if the free variable $Y$ is also the tail variable, and is open if the free variable does not occur in the body atoms. A rule is abstract if it contains no constants. Otherwise, it is instantiated. 

In this work, we use the following types of rules to make up our language bias:
\begin{align*}
	\text{\textbf{Template}}: & r_t(X,Y) \leftarrow r_1(X,V_0),...,r_n(V_n, V_{n+1})\\
	\text{\textbf{HAR}}: & r_t(X,e_k) \leftarrow r_1(X,V_0),...,r_n(V_n, V_{n+1})\\
	\text{\textbf{BAR}}: & r_t(X,e_i) \leftarrow r_1(X,V_0),...,r_n(V_n, e_j)\\
	\text{\textbf{CAR}}: & r_t(X,Y) \leftarrow r_1(X,V_0),...,r_n(V_n, Y)
\end{align*}
where a template is an open (acyclic) abstract rule; a Head Anchored Rule (HAR) is a specialization of a template where the free variable is substituted with a constant; a Both Anchored Rule (BAR) is a specialization of a HAR where the tail variable is replaced with a constant, and a CAR, as introduced in previous sections, is a closed (cyclic) abstract rule. Collectively, abstract rules include CARs and templates, and instantiated rules include HARs and BARs. In particular, templates are used as intermediate rules for generating HARs and BARs only, and will not be included in the learned rule set for inference. This is because templates as rules are too general to differentiate predictions. A HAR characterizes potential candidates in relation of $r_t$ to an entity $e_k$ by a pattern, whereas a BAR highlights a pattern involving the correlation between entities $e_i$ and $e_j$. The CAR is a base rule type included in language biases employed by most of the existing works in that it is often small in size but provides a good base predictive performance.

The reason for selecting these rule types is based on the assumptions of concept stratification and deconstruction. Concept stratification assumes that the learning targets usually have different explanatory complexities, thus rule types of different complexities should be included in the rule space to adapt different targets. For instance, given a correct prediction $r_t(e_0, e_1)$ and an incorrect one $r_t(e_2, e_1)$, both are suggested by the HAR:
\begin{align*}
	& r_t(X,e_1) \leftarrow r_1(X, V_0)
\end{align*}
with confidence $\alpha_1$. As they are suggested with the same confidence, the system can not distinguish one from another. This is a case where the rule space is too general for the learning target. When we allow the more specific BAR in the rule space, and we know that the BAR:
\begin{align*}
	& r_t(X,e_1) \leftarrow r_1(X, e_3)
\end{align*} 
predicts $r_t(e_0, e_1)$ with confidence $\alpha_2 \neq \alpha_1$, the system then can treat the predictions differently based on their confidence. The inclusion of both HAR and BAR is an attempt to stratify the concepts that can be expressed by the system for better adaptivity.

Concept deconstruction assumes a complex concept can be expressed by the combination of simple concepts. For instance, a complex rule that has more than one constants in its body atoms is as follow:
\begin{align*}
	& r_t(X_1,e) \leftarrow r_1(X_1, e_0), r_2(e_0, e_1)
\end{align*}  
and it can be expressed by the conjunction of BARs:
\begin{align*}
	& r_t(X_2,e) \leftarrow r_1(X_2, e_0)\\
	& r_t(X_3,e) \leftarrow r_1(X_3, V_0), r_2(V_0, e_1)
\end{align*}
in that $X_1$ has the same domain as $X_2 \cap X_3$.

\subsection{Algorithm Overview}

In this section, we introduce the GPFL algorithm and discuss the design idea in detail. Above all, GPFL is designed to be a discriminative learner that mines rules for one target predicate at a time. We denote by $r_t \in \mathcal{R}$ the selected learning target, and $I$ the set of positive instances of $r_t$ in a given KG $\mathcal{G}$. As shown in Algorithm.\ref{alg.1} that generates rules for a target $r_t$, GPFL starts by initializing the rule set $F$, and then by calling the \texttt{Generalization} procedure, a rule frequency map $M$ is returned. Map $M$ stores key-value pairs where the key is an abstract rule and the corresponding value is the occurrence of the rule counted during the generalization. In our design, we allow the use of time and space constraints to terminate the system prematurely for accommodating tasks with diverse requirements. Therefore, it is important to sort the abstract rules in $M$ in order to make rules that are more likely to be frequent patterns visited first in the specialization loop. The \texttt{Sort} procedure resolves this by first dividing rules into CARs and instantiated rules of different lengths, then sorting rules in divisions by frequency in descending order, and eventually assemble the sorted divisions into a list $L$ by adding CAR division first and then divisions of instantiated rules with increasing length.

\begin{algorithm}[tb] 
\SetKwInOut{Input}{Input}\SetKwInOut{Output}{Output}
\SetKwFunction{Generalization}{Generalization}
\SetKwFunction{Sort}{Sort}
\SetKwFunction{Ground}{Ground}
\SetKwFunction{Score}{Score}
\SetKwFunction{Specialization}{Specialization}
\SetKwFunction{Constraints}{Constraints}
\SetKwFunction{Quality}{Quality}

\Input{$\mathcal{G}, I, sat, bs, len$}
\Output{learned rule set $F$}

Initialize empty set $F$\;
$M \leftarrow$ \Generalization{$\mathcal{G}, I, sat, bs, len$}\;
$L \leftarrow$ \Sort{M}\;
\For{$l \in L$}{
    $G \leftarrow$ \Ground{$\mathcal{G}, l$}\;
    \eIf{$l$ is a CAR}{
        \Score{$l,G$}\;
        $F \leftarrow F \cup l$\;
    }{
        $S \leftarrow$ \Specialization{$l, G, I$}\;
        \For{$s \in S$}{
            \Score{$s,G$}\;
            $F \leftarrow F \cup s$\;
        }
    }
    \If{\Constraints{}}{
        \textbf{Break}\;
    }
}
$F \leftarrow$ \Quality{$F$}\;
\textbf{return} $F$\;
\caption{Rule Generation for a Target Predicate}
\label{alg.1}
\end{algorithm}

For each abstract rule $l \in L$, GPFL grounds it over $\mathcal{G}$ to produce groundings $G$. Groundings can be used to score the abstract rule $l$ if it is of type CAR, or to derive and evaluate instantiated rules. We define a scoring procedure \texttt{Score} that measures the quality of rules. Various rule quality measures have been proposed in existing works. In this work, we utilize three popular measures, namely the standard confidence (SC) \cite{Galarraga2015}, the smooth confidence (SMC) \cite{Meilicke2019} and the Partial Completeness Assumption confidence (PCA) \cite{Galarraga2015}, for rule evaluation. The support, denoted by $SP_{train}$, of a rule is the number of correct predictions the rule suggests
over the training set, and the body grounding, denoted by $BG_{train}$, is the number of possible groundings of the body atoms of the rule. Now, for a given rule $l$, the standard confidence is defined as:
\begin{equation} \label{eq.1}
    SC(l) = \frac{SP_{train}}{BG_{train}}
\end{equation}
and smooth confidence as:
\begin{equation} \label{eq.2}
    SMC(l) = \frac{SP_{train}}{\eta + BG_{train}}
\end{equation}
where $\eta$ is an offset used to cope with the bias which assigns high confidence to rules that only make a few predictions. In contrast to SC and SMC that operate under the Closed World Assumption (CWA), PCA assumes functionality in target predicates. In particular, given a target $r_t$, for every $i$ such that $r_t(e_i,e_j) \in \mathcal{T}$, PCA only treats triples $r_t(e_i,e_k) \not \in \mathcal{T}$ that contradict the functionality of $r_t$ as negative instances, which is denoted by $FBG_{train}$. Therefore, PCA is defined as:
\begin{equation} \label{eq.3}
    PCA(l) = \frac{SP_{train}}{FBG_{train}}
\end{equation}
In addition, we also define the head coverage (HC) as:
\begin{equation} \label{eq.4}
    HC(l) = \frac{SP_{train}}{|r_t|}
\end{equation}
where $r_t$ is the head atom of $l$, and $|r_t|$ is the number of positive instances of $r_t$.

From line 10 to 14 in Algorithm.\ref{alg.1}, GPFL specializes a template $l$ with distinct constants occurred in $G$ and $I$ into HARs and BARs in $S$, and scores each instantiated rule $s \in S$ with $G$ which is the groundings of $l$ instead of that of $s$ itself. In such a way, GPFL evaluates instantiated rules collectively without the need of grounding each instantiated rule individually. At the end of the loop, the system checks whether the time and space constraints are met or not. If \texttt{Constraints} returns true, the learning terminates, and the quality check procedure \texttt{Quality} is triggered to filter out rules with measures below the pre-defined threshold of selected quality measure, support and head coverage before returning $F$ as the final learned rule set.

\begin{algorithm}[tb] 
\SetKwInOut{Input}{Input}\SetKwInOut{Output}{Output}
\SetKwRepeat{Do}{do}{while}
\SetKwFunction{PathSampler}{PathSampler}
\SetKwFunction{Abstraction}{Abstraction}
\SetKwFunction{Mod}{mod}

\Input{$\mathcal{G}, I, sat, bs, len$}
\Output{rule frequency map $M$}

Initialize empty set $T$ and map $M$\;
$c, sat' \leftarrow 0$\;
\Do{$sat' < sat$}{
    $i \leftarrow$ randomly sample an instance from $I$\;
    $P \leftarrow$ \PathSampler{$\mathcal{G}, i, len$}\;
    \For{$p \in P$}{
        $c \leftarrow c + 1$\;
        $t \leftarrow$ \Abstraction{$p$}\;
        $T \leftarrow T \cup t$\;
        Update $M$ with $p$\;
        \If{\Mod{$c,bs$} = 0}{
            $sat' \leftarrow \frac{|M.keys \cap T|}{|T|}$\;
            $T \leftarrow \emptyset$\;
            \If{$sat' > sat$}{
                \textbf{Break}\;
            }
        }
    }
}
\textbf{Return} $M$\;
\caption{Generalization}
\label{alg.2}
\end{algorithm}

\subsection{Rule Generation}

Rule generation in GPFL is divided into two parts, the generalization part for abstract rules and the specialization part for instantiated rules. Algorithm.\ref{alg.2} details the \texttt{Generalization} procedure. It takes the KG $\mathcal{G}$, positive instance $I$, saturation threshold $sat$, batch size $bs$ and maximum length of rules $len$ as inputs, and produces the rule frequency map $M$. GPFL randomly samples an instance $i \in I$ such that the entities in $i$ are treated as starting nodes, and asks \texttt{PathSampler} to traverse the $len$-hop neighbourhood of $i$ to sample paths $P$. As GPFL is designed as an in-disk system, that is unlike AnyBURL, GPFL does not load the complete KG in memory, instead it requests a graph database that hosts $\mathcal{G}$ to return the neighbourhood of $i$ during traversal. Therefore, contrary to AnyBURL that samples one path at a time, GPFL employs a random walker to sample many paths originated from the starting nodes in an invocation of the \texttt{PathSampler} to balance the number of database queries and the number of retrieved paths. Each path $p \in P$ is turned into an abstract rule $t$ by the procedure \texttt{Abstraction} in which all of the constants in $p$ are replaced with distinct variables. The abstract rule $t$ is then added to the current batch $T$ and used to update map $M$ by logging $t$ as its key and the occurrence of $t$ as its value. Therefore, the known rules in current batch $T$ is the intersection of $T$ and rules in $M$. When the path count $c$ is a multiple of the pre-defined batch size $bs$, current saturation $sat'$ is updated with the ratio of the known rules to all rules in current batch, and if $sat'$ is greater than the saturation threshold $sat$, the \texttt{Generalization} procedure terminates. 

The \texttt{Specialization} procedure in Algorithm.\ref{alg.1} takes a template $l$, groundings $G$ and instances $I$ to produce $S$, a set of instantiated rules derived from $l$. Consider we have a template:
\begin{align*}
    & f_3 = r_t(X, Y) \leftarrow r_1(Y, V_0), r_2(V_0, V_1)
\end{align*}
where $X$ is the free variable, $Y$ the original variable, and $V_1$ the tail variable, and also positive instances of $r_t$:
\begin{align*}
    & I_0 = \{(e_0, e_1), (e_0, e_2), (e_1, e_3)\}
\end{align*}
where the set of substitutions of the free variable is $\{e_0, e_1\}$, we can derive HARs based on $f_3$ and $I_0$ by substituting the free variable with constants. For instance, a possible HAR that can be derived from $f_3$ is: 
\begin{align*}
    & f_4 = r_t(e_0, Y) \leftarrow r_1(Y, V_0), r_2(V_0, V_1)
\end{align*}
where the free variable is replaced by $e_0$. For convenience, we call constants that ground the free variable as free constants, constants grounding original variable the original constants, and constants grounding tail variable the tail constants accordingly.

We compactly represent the groundings of rules as original and tail constants pairs in following discussions. For instance, we can extract compact grounding $(e_1, e_4)$ from a full grounding of $f_3$:
\begin{align*}
    & r_t(e_0, e_1) \leftarrow r_1(e_1, e_2), r_2(e_2, e_4)
\end{align*}
in that $e_1$ and $e_4$ are the original and tail constants in the grounding, respectively. Consider we are given all possible groundings of $f_3$ as:
\begin{align*}
    & G_0 = \{(e_1, e_4), (e_2, e_3), (e_3, e_5)\}
\end{align*}
by joining $G_0$ and $I_0$ by original constants, we can effectively avoid the creation of groundless BARs. For instance, by joining instance $(e_0, e_1)$ and grounding $(e_1, e_4)$ by $e_1$, we have a new pair $(e_0, e_4)$ where $e_0$ is a free constant and $e_4$ is a tail constant. By substituting $f_3$ with $(e_0, e_4)$, that is replacing the free variable $X$ with $e_0$ and the tail variable with $e_4$, we have a BAR derived from $f_3$ as:
\begin{align*}
    & f_5 = r_t(e_0, Y) \leftarrow r_1(Y, V_0), r_2(V_0, e_4)
\end{align*}
which is guaranteed to have at least one grounding. When the \texttt{Specialization} procedure finishes, GPFL will start evaluating the generated instantiated rules collectively.

\subsection{Collective Rule Evaluation} 

We observe that instantiated rules derived from the same templates share the same sequence of predicates. This structural similarity introduced by deductive dependency implies that the groundings of instantiated rules derived from the same templates are either same as or subset of the groundings of the deriving templates. For instance, given the groundings $G_0$, HAR $f_4$ and BAR $f_5$ from previous example, and a new BAR:
\begin{align*}
    & f_6 = r_t(e_1, Y) \leftarrow r_1(Y, V_0), r_2(V_0, e_5)
\end{align*}
the groundings of $f_4$ is $G_0$, that of $f_5$ is $\{(e_1, e_4)\} \in G_0$, and that of $f_6$ is $\{(e_3, e_5)\} \in G_0$. Therefore, instead of invoking the expensive grounding procedure on every instantiated rule, GPFL grounds the templates and uses the groundings of the templates to evaluate instantiated rules collectively. In such a way, GPFL substantially reduces the number of invocations of grounding procedure on large KGs for better efficiency.

\begin{figure}
\centerline{\includegraphics[scale=0.4]{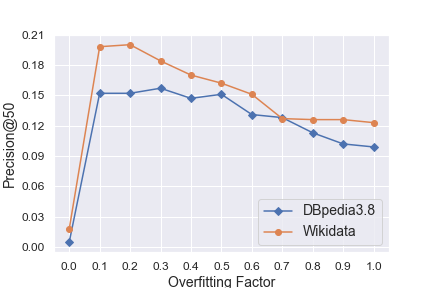}}
\caption{Global average precision of top-50 rules over overfitting factors.}
\label{fig.2}
\end{figure}

\begin{table}[!h]
\centering
\begin{tabular}{lrrrrr}
\toprule
Data  & \#Entities & \#Relationships & \#Types\\
\midrule
DBpedia3.8 & 2.20M & 11.02M & 650\\
Wikidata & 4.00M & 8.40M & 430\\
FB15K-237 & 14.54K & 310K & 237\\
WN18RR & 40.94K & 93K & 11\\
\bottomrule
\end{tabular}
\caption{Statistics of the benchmark datasets.}
\label{tab.1}
\end{table}

\section{Experiments}
In this section, we establish the effectiveness of GPFL through empirical studies over public benchmark datasets on various tasks. By comparing the rules produced by GPFL and AnyBURL, we demonstrate that GPFL learns much more quality rules than AnyBURL in a fixed time frame. Through evaluating rules over KGs of different sizes, we show that the collective approach implemented in GPFL significantly reduces the runtime on evaluating instantiated rules over the baseline approach adopted by existing works. We formally define the overfitting rule, and study the characteristics and effects of overfitting rules through carefully designed experiments with GPFL. By filtering out overfitting rules with a simple validation method, considerable improvements on the predictive performance of learned rules are observed. At last, we report that GPFL performs competitively, with and without validation applied, on knowledge graph completion (KGC) task in comparison to state-of-the-art logic-based and embedding-based methods. 

\subsection{Datasets}
We select four publicly available benchmark datasets, including FB15K-237 \cite{Toutanova2015}, WN18RR \cite{Dettmers2018}, DBpedia3.8 and Wikidata \cite{Galarraga2015}, for experimental evaluations. The statistics of these datasets are reported in Table.\ref{tab.1}. FB15K-237 and WN18RR are popular benchmarks for evaluation on KGC task. Both FB15K-237 and WN18RR are modified versions of the original datasets proposed in \cite{Bordes2013} to mitigate test set leakage introduced by the reverse of test triples being present in the training set. DBpedia3.8 and Wikidata are often considered as large KGs for evaluating the scalability and rule mining capability of rule learning systems. 

\subsection{Implementation}
Unlike most of the existing works that are either implemented in memory or optimized on relational databases, GPFL is implemented in Java on top of the Neo4j\footnote{\url{https://github.com/neo4j/neo4j}} graph database. GPFL uses the core Neo4j API to traverse graph databases for path sampling and rule grounding. All experiments are conducted on AWS EC2 instances that have 8 CPU cores and 64GB RAM. GPFL allows flexible control on scalability by adjusting hyper-parameters to accommodate the specification of running machines. For scalability-related parameters, we set them to values that push running machines to limit. We have made GPFL publicly available at \url{https://github.com/irokin/GPFL}.

\begin{figure}
\centerline{\includegraphics[scale=0.4]{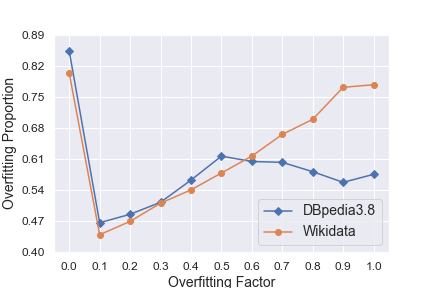}}
\caption{Overfitting proportions over overfitting factors.}
\label{fig.3}
\end{figure}

\begin{table*}[!h]
\centering
\scalebox{0.8}{
\begin{tabular}{llrrrrcrrrr}
\toprule
 & & \multicolumn{4}{c}{DBpedia3.8} & \phantom{a} & \multicolumn{4}{c}{Wikidata}\\
\cmidrule{3-6} \cmidrule{8-11}
Quality & System & All & len=1 & len=2 & len=3 && All & len=1 & len=2 & len=3\\
\midrule
\multirow{2}{*}{All} & GPFL & 3.97M & 59.89K & 1.36M & 2.45M && 563K & 18.21K & 264K & 280K\\
& AnyBURL & 269K & 44.61K & 224K & 0 && 163K & 21.92K & 141K & 0\\
\midrule
\multirow{2}{*}{High} & GPFL & 2.97M & 22.38K & 691K & 2.25M && 271K & 8.94K & 133K & 129K\\
& AnyBURL & 83.31K & 13.79K & 69.51K & 0 && 55.34K & 9.39K & 45.95K & 0\\
\midrule
\multirow{2}{*}{Extreme} & GPFL & 10.83K & 715 & 9.57K & 554 && 12.04K & 739 & 5.52K & 5.77K\\
& AnyBURL & 1.46K & 137 & 1.33K & 0 && 710 & 25 & 685 & 0\\
\bottomrule
\end{tabular}}
\caption{Sizes of learned rules grouped by length ($len$) and quality level.}
\label{tab.2}
\end{table*}

\begin{table}[!h]
\centering
\scalebox{0.82}{
\begin{tabular}{lrrcrrcrr}
\toprule
 & \multicolumn{2}{c}{WN18RR} & \phantom{a} & \multicolumn{2}{c}{FB15K-237} & \phantom{a} & \multicolumn{2}{c}{DBpedia3.8}\\
\cmidrule{2-3} \cmidrule{5-6} \cmidrule{8-9}
Type & Baseline & GPFL && Baseline & GPFL && Baseline & GPFL\\
\midrule
 All & 195.89 & 3.63 && 703.16 & 9.45 && 1285.62 & 69.41\\
 CAR & 0.84 & 0.84 && 2.99 & 2.81 && 21.54 & 21.49\\
 len=1 & 11.09 & 0.36 && 52.63 & 0.56 && 499.91 & 13.27\\
 len=2 & 33.09 & 0.54 && 123.47 & 0.61 && 537.36 & 15.44\\
 len=3 & 150.85 & 1.87 && 524.05 & 5.46 && 294.48 & 22.84\\
\bottomrule
\end{tabular}
}
\caption{Runtimes of rule evaluation approaches measured in various rule groups and reported in seconds.}
\label{tab.3}
\end{table}

\subsection{Rule Mining}
In this experiment, we demonstrate that GPFL mines much more quality rules than AnyBURL in a fixed time frame. We use the smooth confidence with $\eta = 5$ as the quality measure, and set confidence threshold $conf$ to 0.001, support threshold $supp$ to 3 and head coverage threshold $hc$ to 0.001 in the \texttt{Quality} procedure. We follow the method used in \cite{Galarraga2015, Omran2018} to classify rules into high quality rules ($conf \geq 0.1$ and $hc \geq 0.01$), and extremely high quality rules ($conf \geq 0.7$). We run both GPFL and AnyBRUL for 15000s with 6 threads, and report the sizes of rules grouped by different lengths and quality levels. As observed in Table.\ref{tab.2}, GPFL learns more than 10 times the amount of rules produced by AnyBURL on DBpeida, and discovers much more high quality (High) and extremely high quality (Extreme) rules as well. It is worth noting that AnyBURL fails to produce any rules of length 3 because it had trouble saturating rules of length 2 on both datasets, whereas GPFL succeeded in satisfying the saturation threshold and proceeded to generate rules of all lengths.

\subsection{Rule Evaluation Efficiency}

In this experiment, we demonstrate the superiority of the collective approach utilized by GPFL in evaluating instantiated rules over the baseline approach that grounds every mined rule for evaluation. For fair comparison, we implement the baseline approach on Neo4j graph database as well. We select WN18RR, FB15K-237 and DBpedia3.8 to account for the effect of size of KG on the experiment results. For FB15K-237 and DBpedia3.8, we randomly select 20 targets for evaluation, and for WN18RR, we evaluate on all 11 relationship types. The rule set for each benchmark dataset is prepared beforehand, and are divided into CARs and instantiated rules of different lengths (e.g., $len=1$) for fine-grained evaluation. To ensure experiments can be finished in a reasonable time, we allow the evaluation of each target to run for at most 30 minutes. We report the average runtime over all evaluated targets. As shown in Table.\ref{tab.3}, GPFL runs significantly faster than the baseline on all testing datasets when evaluating instantiated rules. For instance, it took GPFL 1.87s to evaluate the same set of instantiated rules of length 3 that took the baseline 150.85s to run on WN18RR. When evaluating CARs, the collective approach in GPFL is reduced to the baseline approach, and thus we observe similar performances between GPFL and the baseline.

\subsection{Overfitting Analysis}
In this section, we formally define the overfitting rule, study the presence and effect of overfitting rules with GPFL, and report our observations. To understand the motivation of investigating overfitting rules, we first define metrics reflecting the predictive performance of learned rules. The test precision of a rule $f$ is defined as:
\begin{equation}
    P_{test}(f) = \frac{SP_{test}}{BG_{test}}
\end{equation}
where $SP_{test}$ is the support and $BG_{test}$ the body groundings of $f$ over the test set. Given a set of target predicates, we use the average of average test precision of top-$k$ rules of each target over all targets as the performance indicator, where the rules are sorted by a selected quality measure. For simplicity, we name this metric the global average precision. Similarly, we can also measure the global average quality over target predicates. As shown in Table.\ref{tab.5}, we report the global average precision and quality over top-$k$ rules on benchmark datasets. Counter-intuitively, by removing certain high-confidence rules via a validation method, we observe consistent improvements on precision on both datasets even though the quality drops dramatically. We argue this phenomenon is partially caused by the presence of overfitting rules, and reason this argument through experiments.

Similar to overfitting models that over-perform on training set yet under-perform on test set, overfitting rules share high quality measured on training set yet have low test precision. Therefore, we consider a rule overfitting if its test precision is smaller than 10\% of its quality. The choice of 10\% is based on our experimental observations. To identify and remove overfitting rules, a simple solution is to measure the precision of rules on a validation set and filter out rules that have a validation precision smaller than $\theta$ percent of the quality. We name $\theta$ the overfitting factor in this simple validation method, and set it to 0.1 in following experiments. We select DBpedia3.8 and Wikidata for experiment where we randomly select 20 targets from each dataset. We select the top 6000 rules from each target to create a collection of top rules, and conduct overfitting analysis on this rule collection.

\begin{table*}[!h]
\centering
\scalebox{0.78}{
\begin{tabular}{llrrcrrrcrrrcrrrcrrr}
\toprule
& & \multicolumn{18}{c}{DBpedia3.8}\\
\cmidrule{3-20}
& & \multicolumn{2}{c}{All} & \phantom{a} & \multicolumn{3}{c}{CAR} & \phantom{a} & \multicolumn{3}{c}{len=1} & \phantom{a} & \multicolumn{3}{c}{len=2} & \phantom{a} & \multicolumn{3}{c}{len=3}\\
\cmidrule{3-4} \cmidrule{6-8} \cmidrule{10-12} \cmidrule{14-16} \cmidrule{18-20}
Measure & Validation & Rules & $ORP_{all}$ && $RP_{all}$ & $ORP_{or}$ & $ORP_{type}$ && $RP_{all}$ & $ORP_{or}$ & $ORP_{type}$ && $RP_{all}$ & $ORP_{or}$ & $ORP_{type}$ && $RP_{all}$ & $ORP_{or}$ & $ORP_{type}$\\
\midrule
\multirow{2}{*}{Standard} & Yes & 6.03K & 0.611 && 0.168 & 0.111 & 0.406 && 0.089 & 0.080 & 0.548 && 0.743 & 0.810 & 0.666 && 0 & 0 & 0\\
& No & 96.29K & 0.915 && 0.019 & 0.011 & 0.519 && 0.082 & 0.083 & 0.922 && 0.841 & 0.858 & 0.934 && 0.058 & 0.049 & 0.780\\
\midrule
\multirow{2}{*}{Smooth} & Yes & 11.06K & 0.499 && 0.096 & 0.074 & 0.386 && 0.057 & 0.060 & 0.526 && 0.608 & 0.719 & 0.590 && 0.239 & 0.144 & 0.302\\
& No & 98.91K & 0.860 && 0.019 & 0.012 & 0.521 && 0.077 & 0.081 & 0.906 && 0.823 & 0.864 & 0.903 && 0.081 & 0.043 & 0.457\\
\midrule
\multirow{2}{*}{PCA} & Yes & 3.34K & 0.736 && 0.071 & 0.053 & 0.548 && 0.067 & 0.054 & 0.597 && 0.862 & 0.893 & 0.762 && 0 & 0 & 0\\
& No & 90.54K & 0.952 && 0.014 & 0.013 & 0.837 && 0.091 & 0.091 & 0.953 && 0.870 & 0.877 & 0.960 && 0.024 & 0.017 & 0.653\\

\midrule
\midrule

& & \multicolumn{18}{c}{Wikidata}\\
\cmidrule{3-20}
& &  \multicolumn{2}{c}{All} & \phantom{a} & \multicolumn{3}{c}{CAR} & \phantom{a} & \multicolumn{3}{c}{len=1} & \phantom{a} & \multicolumn{3}{c}{len=2} & \phantom{a} & \multicolumn{3}{c}{len=3}\\
\cmidrule{3-4} \cmidrule{6-8} \cmidrule{10-12} \cmidrule{14-16} \cmidrule{18-20}
Measure & Validation & Rules & $ORP_{all}$ && $RP_{all}$ & $ORP_{or}$ & $ORP_{type}$ && $RP_{all}$ & $ORP_{or}$ & $ORP_{type}$ && $RP_{all}$ & $ORP_{or}$ & $ORP_{type}$ && $RP_{all}$ & $ORP_{or}$ & $ORP_{type}$\\
\midrule
\multirow{2}{*}{Standard} & Yes & 9.14K & 0.402 && 0.038 & 0.015 & 0.159 && 0.030 & 0.035 & 0.469 && 0.429 & 0.557 & 0.522 && 0.503 & 0.557 & 0.445\\
& No & 78.03K & 0.852 && 0.007 & 0.002 & 0.286 && 0.061 & 0.067 & 0.934 && 0.584 & 0.619 & 0.903 && 0.348 & 0.313 & 0.767\\
\midrule
\multirow{2}{*}{Smooth} & Yes & 12.65K & 0.439 && 0.040 & 0.011 & 0.126 && 0.031 & 0.039 & 0.545 && 0.508 & 0.626 & 0.541 && 0.421 & 0.326 & 0.340\\
& No & 84.22K & 0.803 && 0.008 & 0.002 & 0.236 && 0.057 & 0.064 & 0.898 && 0.557 & 0.583 & 0.840 && 0.378 & 0.351 & 0.747\\
\midrule
\multirow{2}{*}{PCA} & Yes & 5.36K & 0.563 && 0.024 & 0.009 & 0.209 && 0.037 & 0.036 & 0.541 && 0.564 & 0.575 & 0.574 && 0.375 & 0.380 & 0.571\\
& No & 78.02K & 0.917 && 0.006 & 0.004 & 0.713 && 0.061 & 0.063 & 0.949 && 0.583 & 0.590 & 0.928 && 0.350 & 0.342 & 0.897\\
\bottomrule
\end{tabular}}
\caption{The experiment results for overfitting analysis.}
\label{tab.4}
\end{table*}

\begin{table}[!h]
\centering
\scalebox{0.82}{
\begin{tabular}{llrrcrr}
\toprule
 & & \multicolumn{2}{c}{DBpedia3.8} & \phantom{a} & \multicolumn{2}{c}{Wikidata}\\
\cmidrule{3-4} \cmidrule{6-7}
Top-k & Validation & Precision & Quality && Precision & Quality\\
\midrule
\multirow{2}{*}{5} & Yes & 0.163 & 0.649 && 0.26 & 0.605\\
& No & 0.023 & 0.914 && 0.05 & 0.918\\
\midrule
\multirow{2}{*}{10} & Yes & 0.183 & 0.618 && 0.247 & 0.586\\
& No & 0.011 & 0.907 && 0.045 & 0.912\\
\midrule
\multirow{2}{*}{20} & Yes & 0.184 & 0.584 && 0.237 & 0.592\\
& No & 0.012 & 0.897 && 0.038 & 0.901\\
\midrule
\multirow{2}{*}{50} & Yes & 0.152 & 0.536 && 0.198 & 0.553\\
& No & 0.005 & 0.879 && 0.018 & 0.881\\
\midrule
\multirow{2}{*}{100} & Yes & 0.144 & 0.511 && 0.17 & 0.543\\
& No & 0.004 & 0.859 && 0.015 & 0.855\\
\bottomrule
\end{tabular}}
\caption{Global average precision and quality over top-$k$ rules on DBpedia3.8 and Wikidata with and without validation applied.}
\label{tab.5}
\end{table}

We set out to answer following questions: 1.) which types of rules make up the largest portion of overfitting rules; 2.) which types of rules are more likely to be overfitting; 3.) the effect of different choices of quality measure on overfitting rules; 4.) the effectiveness of the validation method at removing overfitting rules, 5.) the effect of removal of overfitting rules on the predictive performance. To answer these questions, we need to define a set of terms. We consider the overfitting rule proportion, denoted by $ORP_{s}(t)$, as the proportion of overfitting rules of type $t$ to a rule space $s$. The domain of $t$ includes "All" as in all rule types, CAR and instantiated rule of different lengths, and that of $s$ allows "all" as in space of all learned rules, "or" as in space of all overfitting rules, and "type" as in space of all rules of type $t$. For instance in Table.\ref{tab.4}, under the standard measure and with validation on Wikidata, the $ORP_{type}(len=1)$, which is the proportion of overfitting rules of instantiated rule of length 1 (as $t$ is $len=1$) to all instantiated rules of length 1 (as $s$ is $type$), is 0.469. In other words, $ORP_{type}(len=1)$ indicates that the probability of an instantiated rule of length 1 being overfitting is 46.9\%. Similarly, we define the rule proportion, denoted by $RP_{all}(t)$, as the proportion of all rules of type $t$ to all learned rules. Again in Table.\ref{tab.4}, under the standard measure and with validation on Wikidata, the $RP_{all}(CAR)$ is 0.038, which translates as CARs make up 3.8\% of all rules. We name $ORP_{all}(All)$ the overfitting proportion as it states the overfitting rule to all rule ratio.

\begin{table*}[!h]
\centering
\scalebox{0.8}{
\begin{tabular}{lrrrrcrrrr}
\toprule
 & \multicolumn{4}{c}{FB15K-237} & \phantom{a} & \multicolumn{4}{c}{WN18RR}\\
\cmidrule{2-5} \cmidrule{7-10}
Algorithm & MRR & Hits@10 & Hits@3 & Hits@1 && MRR & Hits@10 & Hits@3 & Hits@1\\
\midrule
DistMult \cite{Yang2015} & 0.241 & 0.419 & 0.263 & 0.155 && 0.430 & 0.490 & 0.440 & 0.390\\
ComplEx \cite{Trouillon2016} & 0.247 & 0.428 & 0.275 & 0.158 && 0.440 & 0.510 & 0.460 & 0.410\\
ConvE \cite{Dettmers2018} & 0.316 & 0.491 & 0.350 & 0.239 && 0.460 & 0.480 & 0.430 & 0.390\\
R-GCN+ \cite{Schlichtkrull2018} & 0.249 & 0.417 & 0.264 & 0.151 && - & - & - & -\\
TuckER \cite{Balazevic2019} & 0.358 & 0.544 & 0.394 & 0.266 && 0.470 & 0.526 & 0.482 & \underline{0.443}\\
RotatE \cite{Sun2019} & 0.338 & 0.533 & 0.375 & 0.241 && 0.476 & 0.571 & 0.492 & 0.428\\
QuatE \cite{Zhang2019} & \underline{0.366} & \underline{0.556} & \underline{0.401} & \underline{0.271} && \underline{0.488} & \underline{0.582} & \underline{0.508} & 0.438\\
\midrule
AMIE+ \cite{Galarraga2015} & - & 0.409 & - & 0.174 && - & 0.388 & - & 0.358 \\
Neural LP \cite{Yang2017} & 0.240 & 0.362 & - & - && 0.435 & 0.566 & 0.434 & 0.371\\
RuleN \cite{Meilicke2018} & - & 0.420 & - & 0.182 && - & 0.536 & - & 0.427\\
DRUM \cite{Sadeghian2019} & \underline{0.343} & \underline{0.516} & \underline{0.378} & \underline{0.255} && \underline{0.486} & \underline{0.586} & \underline{0.513} & 0.425\\
AnyBURL \cite{Meilicke2019} & 0.301 & 0.484 & 0.341 & 0.227 && 0.471 & 0.537 & 0.488 & \underline{0.442} \\
\midrule
GPFL-ins0-car3 & 0.253 & 0.421 & 0.285 & 0.189 && 0.455 & 0.529 & 0.475 & 0.423\\
GPFL-ins1-car3 & 0.315 & 0.498 & 0.355 & 0.241 && 0.479 & 0.552 & 0.499 & 0.448\\
GPFL-ins2-car3 & 0.283 & 0.459 & 0.318 & 0.214 && 0.471 & 0.541 & 0.486 & 0.443\\
GPFL-ins3-car3 & 0.277 & 0.448 & 0.311 & 0.209 && 0.453 & 0.499 & 0.465 & 0.433\\
GPFL-Ensemble & \underline{0.322} & \underline{0.504} & \underline{0.362} & \underline{0.247} && \underline{0.480} & \underline{0.552} & \underline{0.500} & \underline{0.449}\\
\bottomrule
\end{tabular}}
\caption{KGC in default setting. The top section contains embedding-based methods; the middle section includes logic-based methods, and the bottom section reports GPFL results under various configurations. Best results in each section are underlined.}
\label{tab.6}
\end{table*}

\begin{table*}[!h]
\centering
\scalebox{0.8}{
\begin{tabular}{lrrrrcrrrr}
\toprule
 & \multicolumn{4}{c}{FB15K-237} & \phantom{a} & \multicolumn{4}{c}{WN18RR}\\
\cmidrule{2-5} \cmidrule{7-10}
Algorithm & MRR & Hits@10 & Hits@3 & Hits@1 && MRR & Hits@10 & Hits@3 & Hits@1\\
\midrule
AnyBURL \cite{Meilicke2019} & 0.269 & 0.452 & 0.311 & 0.193 && 0.288 & 0.345 & 0.302 & \underline{0.263}\\
GPFL-ins3-car3-valid & \underline{0.283} & \underline{0.469} & \underline{0.328} & \underline{0.205} && \underline{0.294} & \underline{0.361} & \underline{0.311} & \underline{0.263}\\
GPFL-ins3-car3 & 0.249 & 0.432 & 0.288 & 0.177 && 0.257 & 0.292 & 0.266 & 0.242\\
\bottomrule
\end{tabular}}
\caption{KGC results in random setting. The "valid" suffix means the algorithm runs with validation applied. Best results are underlined.}
\label{tab.7}
\end{table*}

Now, by analyzing Table.\ref{tab.4} horizontally, we can answer questions 1 and 2. We observe that the $ORP_{or}$, the overfitting rule of certain type to all overfitting rule ratio, of instantiated rules are generally much larger than that of CAR. On Wikidata under the smooth measure and without validation, the $ORP_{or}(len=2)$ is 0.583, and that of $ORP_{or}(len=3)$ is 0.351, which means over 93\% of overfitting rules are long instantiated rules. On DBpedia3.8, the proportion is also over 90\%. Although we can argue the large contribution of overfitting rules made by long instantiated rules is attributed to the observation that $ORP_{or}$ is proportional to $RP_{all}$ and long instantiated rules often have large $RP_{all}$, long instantiated rules still make up the largest portion of overfitting rules. By comparing $ORP_{type}$ under the smooth confidence between CAR and instantiated rules on both datasets, we observe that on DBpedia3.8 without validation, the instantiated rules of all lengths have a 76\% average probability of being overfitting, whereas CAR has 52\%. On Wikidata, the probability of instantiated rules becomes 82\% compared to 23\% of CAR. Considering long instantiated rules are not only larger in size than CARs but also more likely to be overfitting, it is more important for instantiated rule learners to identify and remove overfitting rules than learners that only mine abstract rules.

To answer questions 3 and 4, we analyze Table.\ref{tab.4} vertically. We consider a quality measure better than another if it has smaller overfitting proportion while maintaining a larger rule space. By this criterion, the smooth confidence outperforms standard confidence and PCA in that without validation, smooth confidence has average 91.56K rules and 83\% $ORP_{all}$ over both datasets, whereas standard confidence has 87.16K and 88\%, and PCA has 84.28K and 93\%. The advantage of smooth confidence becomes even more evident when validation is applied, where it has 11.85K and 46\% in comparison to 7.5K and 51\% of standard confidence and 4.3K and 64\% of PCA. One perspective to evaluate the effectiveness of the validation method is to compare the difference in overfitting proportion before and after validation. As shown in Figure.\ref{fig.3}, by changing the overfitting factor from 0 to 0.1, the overfitting proportion drops dramatically, and with increasing factor, it recovers gradually partially because with higher factor, the filter removes more rules yet the overfitting rules that have high validation precision remain untouched, thus the overfitting proportion increases. Correspondingly in Table.\ref{tab.4}, we also observe significant drops in $ORP_{all}$ and $ORP_{type}$ over all comparison pairs. At last, we investigate the effect of removal of overfitting rules on predictive performance. As shown in Figure.\ref{fig.2}, the global average precision of top-50 rules increases from near 0 to around 20\% with overfitting factor being set from 0 to 0.1. In Table.\ref{tab.5}, we also observe significant improvements by having validation applied on both datasets. As without validation, the precision of top rules is extremely bad, we argue that the impact of overfitting rules, especially on instantiated rule learners, is too significant to ignore and must be handled properly. 

In conclusion, we observe that long instantiated rules make up the largest portion of overfitting rules; instantiated rules are more likely to be overfitting than abstract rules; the choice of quality measure considerably affects the overfitting proportion, and the smooth confidence is better than standard confidence and PCA according to our criterion; the validation method can effectively filter out a large portion of overfitting rules, and the predictive performance improves significantly with validation applied.

\subsection{Knowledge Graph Completion}
A knowledge graph completion (KGC) query takes the form of $r_t(e_i, ?)$ or $r_t(?, e_i)$ where $r_t$ is the learning target, and the question mark is expected to be replaced with candidates $e \in \mathcal{E}$ that are suggested by the learned rules such that predictions $r_t(e_i, e)$ or $r_t(e, e_i)$ for target $r_t$ are proposed. We follow the evaluation protocol used in \cite{Bordes2013} to evaluate GPFL on KGC task. GPFL answers both head queries $r_t(?, e)$ and tail queries $r_t(e, ?)$ that are created by corrupting test triples. For reporting the experiment results, we use hits@1, hits@3, hits@10 and mean reciprocal rank (MRR), all in the filtered setting. As a prediction can be suggested by multiple rules, it introduces complexity in ranking the predictions. In this work, we use the maximum aggregation strategy proposed in AnyBURL \cite{Meilicke2019} to rank predictions. In particular, predictions are sorted by the maximum of the confidence of rules suggesting the predictions, and if there are ties among predictions, the tied predictions are sorted by recursively comparing the next highest confidence of suggesting rules until all ties are resolved.

We select FB15K-237 and WN18RR for KGC evaluation. As GPFL allows fine-tuning on composition of learned rule types, we evaluate GPFL over various rule composition configurations. Specifically, the notation "ins$A$-car$B$" is used to indicate the maximal length of instantiated rules as $A$ and that of CARs as $B$. For instance, "ins0-car3" depicts a configuration that only learns CARs of maximum length of 3, whereas GPFL with "ins3-car3" learns both instantiated rules and CARs of maximum length of 3. To respond to the observation where the performance of target predicates differs significantly under different configurations, we also include an ensemble mode that aggregates the best performing rules for different targets from learned rules under various configurations to create an optimal rule space for all target predicates. Table.\ref{tab.6} reports the KGC results in the default setting where we use the default data splits of FB15K-237 and WN18RR downloaded from the source\footnote{\url{http://web.informatik.uni-mannheim.de/AnyBURL/}} to evaluate GPFL for comparability. In contrast, results in Table.\ref{tab.7} are evaluated on a re-split of the original dataset into training/validation/test sets in a 6:2:2 ratio such that the new validation set is much larger than that in the default setting. In Table.\ref{tab.6}, the performance of GPFL-ins1-car3 and GPFL-Ensemble are competitive compared to strong logic-based and embedding-based baselines. In comparison to embedding methods that can only perform transductive inference, GPFL and other logic-based methods are capable of interpretable inductive inference which is crucial to many real-world applications. As observed in Table.\ref{tab.7}, the considerable increase in performance after applying validation to GPFL confirms our expectation that the removal of overfitting rules benefits the predictive performance of learned rules.

\section{Conclusion}
In this paper, we present the GPFL system, a probabilistic rule learner optimized to mine instantiated rules. By utilizing the idea of template for optimizing both rule evaluation and generation, GPFL is capable of efficiently learning more quality instantiated rules than existing works. Through experiments, we reveal the characteristics and impact of overfitting rules and conclude that instantiated rule learners can benefit significantly from the filtering of overfitting rules. Eventually, we show that GPFL has competitive performance on KGC task.

\begin{acks}
We would like to thank Christian Meilicke from University of Mannheim for his valuable opinions and discussions related to this work.
\end{acks}

\bibliographystyle{ACM-Reference-Format}
\bibliography{bibfile}

\end{document}